\documentclass[letterpaper, 10 pt, conference]{ieeeconf}  
\IEEEoverridecommandlockouts                              
\usepackage{epsfig} 
\usepackage{graphicx}
\usepackage[export]{adjustbox} 
\usepackage{amsmath} 
\usepackage{amssymb}  
\usepackage{multirow}
\usepackage{xspace}   
\usepackage{xcolor}   
\usepackage{subcaption}
\usepackage{array}
\usepackage{tabularx}
\usepackage{booktabs}
\usepackage[dvipsnames, table]{xcolor}
\usepackage{pifont} 
\newcommand{\cmark}{\ding{51}} 
\newcommand{\xmark}{\ding{55}} 
\newcolumntype{L}[1]{>{\raggedright\arraybackslash}p{#1}}
\newcolumntype{C}[1]{>{\centering\arraybackslash}p{#1}}
\usepackage{hyperref}

\newcommand{\new}{\vskip 0em}

\newcommand{\name}{\textit{RadarSFD}\xspace} 
\newcommand\website[1]{\textcolor{blue}{#1}}

\title{\LARGE \bf \name: Single-Frame Diffusion with Pretrained Priors for Radar Point Clouds}
\author{Bin Zhao, Nakul Garg \\
\thanks{Bin Zhao and Nakul Garg are with Rice University, Houston TX 77005 USA. Email:\{{\tt\small bz35@rice.edu, nakul@rice.edu}\}.
}}

\begin{document}
\maketitle
\thispagestyle{empty}
\pagestyle{empty}

\begin{abstract}
Millimeter-wave radar provides perception robust to fog, smoke, dust, and low light, making it attractive for size, weight, and power constrained robotic platforms. Current radar imaging methods, however, rely on synthetic aperture or multi-frame aggregation to improve resolution, which is impractical for small aerial, inspection, or wearable systems. We present \name, a conditional latent diffusion framework that reconstructs dense LiDAR-like point clouds from a single radar frame without motion or SAR. Our approach transfers geometric priors from a pretrained monocular depth estimator into the diffusion backbone, anchors them to radar inputs via channel-wise latent concatenation, and regularizes outputs with a dual-space objective combining latent and pixel-space losses. On the RadarHD benchmark, \name achieves state-of-the-art performance against baseline models. Qualitative results show recovery of fine walls and narrow gaps, and experiments across new environments confirm strong generalization. Ablation studies highlight the importance of pretrained initialization, radar BEV conditioning, and the dual-space loss. Together, these results establish the first practical single-frame, no-SAR mmWave radar pipeline for dense point cloud perception in compact robotic systems. The project page is available at \website{\href{https://phi-lab-rice.github.io/RadarSFD/}{https://phi-lab-rice.github.io/RadarSFD/}}
\end{abstract}

\section{Introduction}
Millimeter-wave (mmWave) radar is emerging as a practical alternative to LiDAR for robotic perception. 
Unlike LiDAR, mmWave signals penetrate fog, smoke, dust, and low light, making them robust in conditions where optical sensors struggle \cite{bijelic2018benchmark, hawkeye}. 
This robustness is especially attractive for platforms with tight size, weight, and power (SWaP) budgets, such as autonomous inspection drones in confined spaces, compact robots for search-and-rescue, or wearable devices \cite{sanket2020evdodgenet, garg2023sirius, bai2022spidr, talwekar2022towards}. 
These platforms cannot afford bulky rotating rigs, linear actuators, or multi-view scanning mechanisms. 
Instead, they need compact modules that can deliver single-frame spatial perception without any ego-motion or synthetic aperture. 
In this work, we ask: \textit{can a single radar capture yield a dense, LiDAR-like point cloud suitable for robotic perception?}
\new

\begin{figure}[t]
    \centering
    \setlength{\tabcolsep}{0pt} 
    \renewcommand{\arraystretch}{1.0}
    
    \begin{tabular}{|c|c|c|c|c|} 
    \hline
    \includegraphics[width=0.185\linewidth]{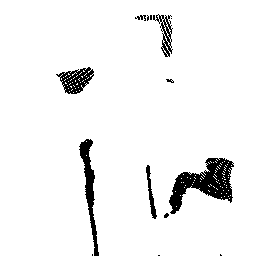} &
    \includegraphics[width=0.185\linewidth]{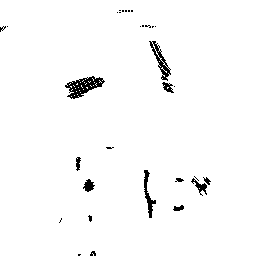} &
    \includegraphics[width=0.185\linewidth]{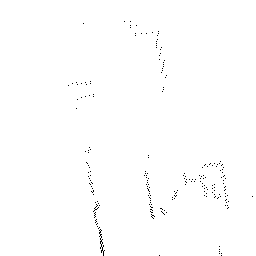} &
    \includegraphics[width=0.185\linewidth]{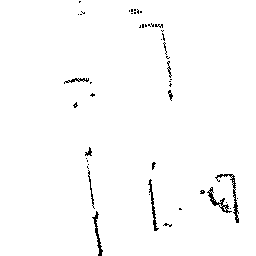} &
    \includegraphics[width=0.185\linewidth]{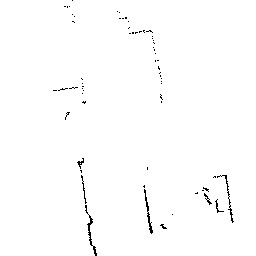} \\ 
    \hline
    \multicolumn{1}{c}{(a)} &
    \multicolumn{1}{c}{(b)} &
    \multicolumn{1}{c}{(c)} &
    \multicolumn{1}{c}{(d)} &
    \multicolumn{1}{c}{(e)} \\ 
    \end{tabular}
    \caption{Estimated point cloud reconstructions:
    (a) RadarHD (41 input frames) \cite{radarhd}, (b) RadarHD single-frame,
    (c) Zhang et al. \cite{ral24}, (d) \name~(Ours), and (e) LiDAR ground truth.}
    \vspace{-0.1in}
    \label{fig-intro}
\end{figure}

Traditional radar imaging faces a fundamental limitation: small apertures result in low angular resolution \cite{spidr_cacm}.  
Classical processing pipelines such as beamforming and CFAR (Constant False Alarm Rate) extract reliable points \cite{cfar}, but the resulting point clouds remain sparse and miss fine structures.  
To improve the resolution, existing approaches use synthetic aperture radar (SAR), which requires carefully controlled ego-motion or temporal stacking of multiple frames.  
Recently, learning-based methods have advanced the field significantly.  
For instance, RadarHD \cite{radarhd} introduced an asymmetric VAE that collects 40 consecutive past frames to produce denser point clouds.  
Luan et al. \cite{icra24} reduces the temporal stacks to 5 frames, while Zhang et al. \cite{ral24} adapts diffusion to 2D (Bird’s eye view) BEV images. 
These are important works that demonstrate the promise of AI-based radar perception.  
However, as we show in Figure~\ref{fig-intro}, when temporal stacks or SAR are removed, RadarHD degrades significantly under the single-frame setting, producing blurred and fragmented outputs.  
The result from Zhang et al., while single-frame, shows visibly coarser resolution compared to LiDAR.  
In contrast, our approach recovers fine walls and gaps from a single frame without sacrificing radar resolution, showing that single-frame, no-SAR radar remains an open and impactful challenge.  
\new

Our approach builds on the intuition that radar alone does not need to learn the entire structure of the world from scratch. 
Instead, we can transfer the structural priors encoded in large pretrained generative models. 
Recent works in monocular depth estimation show that diffusion models like Marigold \cite{marigold} learn strong geometric priors about walls, corners, and object boundaries. 
Inspired by this, we propose to condition a latent diffusion model on the raw radar BEV. 
The pretrained priors act as a “world model”, while the radar input anchors these priors to the actual scene. 
Similar cross-modality transfer has shown success in domains such as underwater restoration \cite{slurpp}, supporting the idea that pretrained diffusion backbones can be adapted across sensing modalities. 
We also build on prior work's insight that lightly thresholded radar BEVs preserve weak reflections and sidelobes that carry useful information \cite{radarhd}. 
Together, these insights form the basis of our method.
\new

\begin{figure*}[htb]
    \centering
    \setlength{\tabcolsep}{0pt} 
    \renewcommand{\arraystretch}{1.0} 
    \vspace{0.05in}
    \begin{tabular}{|c|c|c|c|c|} 
     \hline
     \includegraphics[width=0.308\linewidth, valign=m]{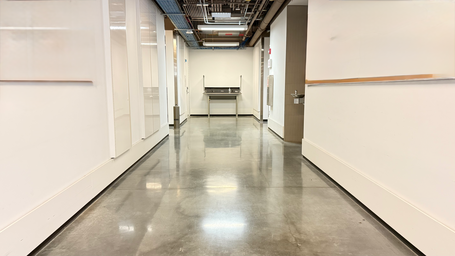} &
     \includegraphics[width=0.173\linewidth, valign=m]{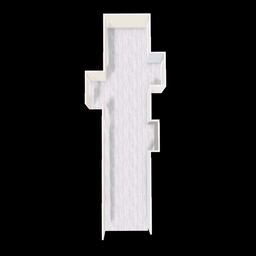} &
     \includegraphics[width=0.173\linewidth, valign=m]{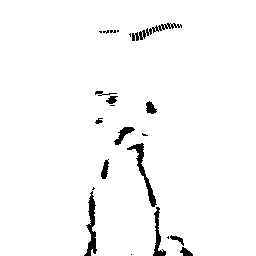} &
     \includegraphics[width=0.173\linewidth, valign=m]{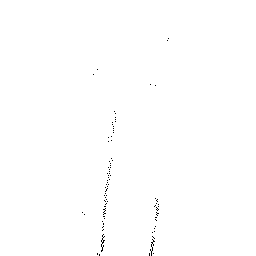} &
     \includegraphics[width=0.173\linewidth, valign=m]{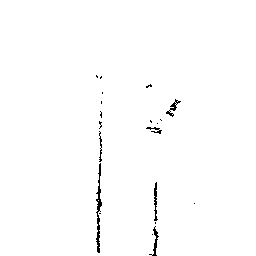} \\ 
     \hline
     \includegraphics[width=0.308\linewidth, valign=m]{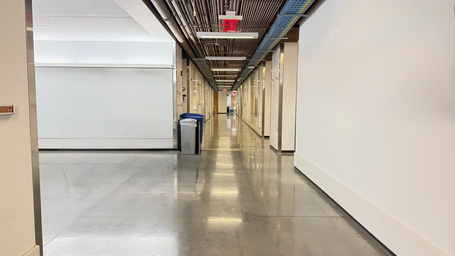} &
     \includegraphics[width=0.173\linewidth, valign=m]{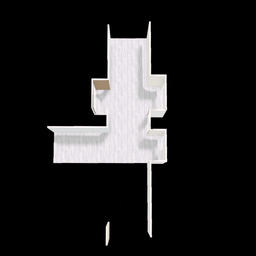} &
     \includegraphics[width=0.173\linewidth, valign=m]{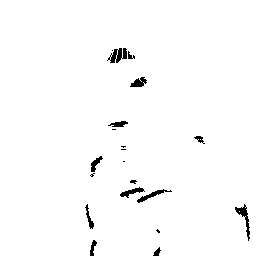} &
     \includegraphics[width=0.173\linewidth, valign=m]{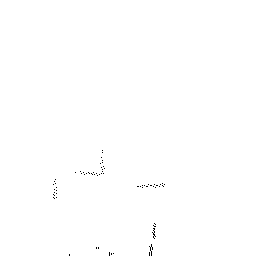} &
     \includegraphics[width=0.173\linewidth, valign=m]{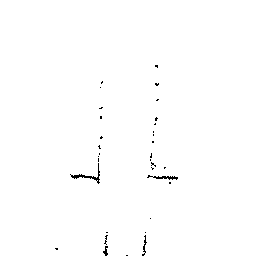} \\ 
     \hline
    \multicolumn{1}{c}{\small (a) RGB image of unseen building} &
    \multicolumn{1}{c}{\small (b) 3D layout BEV } &
    \multicolumn{1}{c}{\small (c) RadarHD \cite{radarhd}} &
    \multicolumn{1}{c}{\small (d) Zhang et al. \cite{ral24}} &
    \multicolumn{1}{c}{\small (e) \name (ours)} \\
    \end{tabular}
    
    \caption{Real world test for generalization with completely unseen data in our campus building. All models are trained on the same radar dataset from RadarHD \cite{radarhd}: (a) RGB image of unseen environment, (b) 3D floor plan layouts, (c) RadarHD single-frame baseline, (d) Zhang et al. diffusion baseline \cite{ral24}, and (e) \name (ours) single-frame latent diffusion.}
    \vspace{-0.2in}
    \label{fig-generalization}
\end{figure*}

We introduce \name (Radar Single-Frame Diffusion), a conditional latent diffusion pipeline for single-frame radar-to-LiDAR translation. 
Our model operates in the latent space of a frozen Stable Diffusion VAE for efficiency. 
We initialize the U-Net from Marigold to inject pretrained depth and shape priors. 
Conditioning is applied by channel-wise concatenation of radar BEV latents and noisy LiDAR BEV latents, aligning the input geometry directly with the generative process. 
To ensure scene faithfulness, we combine the standard latent space MSE loss with pixel-space L1, SSIM \cite{ssim}, and LPIPS \cite{lpips}. This design mitigates the common failure mode of diffusion models -- hallucination/producing plausible but incorrect scenes -- by tethering generations to the specific radar input.
\new

To evaluate, we use Chamfer distance (CD) \cite{barrow1977parametric} and Modified Hausdorff distance (MHD) \cite{dubuisson1994modified}, which measure point-to-point and structural similarity between generated point clouds and the LiDAR ground truth.  
Compared to the single-frame RadarHD baseline, our method reduces CD from 56\,cm to 35\,cm and MHD from 45\,cm to 28\,cm, showing much sharper reconstructions from a single radar frame.  
Our performance is also close to the recent Zhang et al. diffusion method \cite{ral24}, which reports 38\,cm CD and 29\,cm MHD, but importantly that approach achieves its results by lowering range resolution, whereas our method maintains the native 4\,cm resolution.  
Further, even against the multi-frame RadarHD with 41 stacked frames, our method remains competitive and slightly better, with about a 20\% reduction in CD and an 18\% reduction in MHD.  
As Figure~\ref{fig-intro} shows, this enables us to recover fine walls and narrow gaps even in static, no-SAR settings.  
Figure~\ref{fig-generalization} further demonstrates strong generalization across new environments.  
Our ablation studies confirm that pretrained initialization, BEV concatenation, and the dual-space loss are key to achieving this performance.  
 
\new 

In summary, this paper makes three contributions:
\begin{itemize}
    \item A no-SAR radar perception pipeline that generates dense, LiDAR-like point clouds from single-frame radar, suitable for SWaP-constrained platforms.
    \item A conditional latent diffusion model that transfers monocular-depth priors into radar-to-LiDAR translation and achieves state-of-the-art accuracy.
    \item A dual-space objective function and ablation study to make design choices in pretraining, conditioning, and input representation, which provides a practical reproducible recipe for radar-to-LiDAR point clouds.
\end{itemize}

\section{Related Work}
Conventional mmWave radar signal processing pipelines start with transforming the raw I/Q signals into a Range-Doppler-Angle data cube. Standard angle estimation is done via beamforming with an FFT across the antenna array, yet its angular resolution is fundamentally limited by the physical aperture, a constraint known as the Rayleigh limit \cite{hansen2005fundamental}. To surpass this limitation, super-resolution algorithms like MVDR, ESPRIT and MUSIC are often employed, providing much finer angular estimates at the cost of higher computational complexity \cite{mvdr,esprit,music, garg2021owlet}. Separately, the range resolution is constrained by the chirp bandwidth, which can cause objects close in distance to merge into a single detection \cite{garg2025large, yao2023radar}. After the data cube is formed, CFAR algorithms are typically applied to detect targets by dynamically setting thresholds against the local clutter \cite{cfar,socfar,cacfar,gocfar,oscfar}.
\new

To address these resolution limitations, synthetic aperture radar (SAR) techniques are commonly employed \cite{mimo-sar,aditiya,qian20203d}. While SAR significantly improves angular resolution, it requires precise motion control and temporal accumulation. This makes SAR impractical for SWaP-platforms or real-time applications where motion is constrained.
\new

Recent advances in machine learning have opened new avenues in radar perception enhancement. RadarHD \cite{radarhd} pioneered the use of asymmetric variational autoencoders (VAE) to transform sparse radar range-azimuth heatmaps into dense, LiDAR-like representations. RadarHD utilizes lightly thresholded 2D range-azimuth BEV images from range and angle FFT outputs, using temporal stacks of 40 frames as synthetic aperture and dense LiDAR BEV images as supervision. While effective, the VAE architecture produces relatively blurry outputs compared to more advanced generative models. Similar VAE-based approaches have demonstrated deployment feasibility on UAV platforms \cite{radcloud}, though they still rely on multi-frame inputs.
\new

Conditional Generative Adversarial Networks (cGAN) have also been explored for radar perception enhancement, particularly for domain-specific applications. HawkEye \cite{hawkeye} designs a cGAN architecture to recover high-resolution depth maps from conditioning on SAR-aided mmWave heatmaps, specifically for vehicle imaging in fog conditions. Similarly, Mi-Shape \cite{mishape} applies cGAN for human pose estimation from mmWave signals, generating silhouettes and predicting joint locations. While GAN-based approaches can produce sharp and high-quality outputs with adversarial training, these radar cGANs are typically optimized for specific object categories. This domain specificity limits their generalizability compared to more recent generative approaches.
\new

Recent diffusion approaches have advanced radar perception significantly. Diffusion models demonstrate superior denoising and generative capabilities across general-purpose computer vision tasks \cite{ddpm,sd,controlnet}, making them attractive for radar-to-LiDAR translation. The Luan et al. work \cite{icra24} introduced a conditional denoising diffusion probabilistic model (DDPM) that processes 5 temporally-stacked LiDAR BEVs with pixel values representing heights conditioned on the radar BEVs to denoise the corrupted LiDAR BEVs and produce enhanced LiDAR-like radar outputs, employing a weighted-L1 loss to handle sparse regions effectively. The Zhang et al. method \cite{ral24} adopted advanced diffusion formulations including Elucidated Diffusion Models (EDM) \cite{edm} and consistency sampling \cite{cm} to replace iterative DDPM denoising. Their preprocessing pipeline retains traditional range-azimuth FFT processing while incorporating perceptual losses such as LPIPS \cite{lpips}, which is widely employed in computer vision for measuring perceptual similarity. Despite these advances, existing diffusion-based methods depend on temporal stacks or motion cues, and those that achieve single-frame processing sacrifice range and angle resolution. 

Our work addresses this gap by employing a latent diffusion architecture that combines a pretrained VAE encoder-decoder with a denoising U-Net. The VAE encoder maps 2D LiDAR and radar BEV images to latent space, where our proposed U-Net performs denoising operations before the decoder reconstructs enhanced range-azimuth BEV outputs. This approach leverages the superior denoising capabilities of diffusion models while operating on single-frame radar inputs, eliminating the need for temporal aggregation while maintaining full range and image resolution (Table~\ref{tab-related}).

\begin{table}[t]
\vspace{0.05in}
\centering
\caption{Summary of related work}
\label{tab-related}
\setlength{\tabcolsep}{6pt}
\renewcommand{\arraystretch}{1.25}
\begin{tabular}{L{0.25\linewidth} L{0.18\linewidth} L{0.15\linewidth} C{0.15\linewidth}}
\toprule
\textbf{Methods} & \textbf{Arch.} & \textbf{Input} & \textbf{Single-frame (No SAR)} \\
\midrule
RadarHD \cite{radarhd}   & VAE        & 2D BEV          & \xmark \\
RadCloud \cite{radcloud} & VAE        & 2D BEV          & \xmark \\
HawkEye \cite{hawkeye}   & cGAN       & 3D Voxel        & \xmark \\
MiShape \cite{mishape}   & cGAN       & 2D FEV          & \cmark \\
Luan et al. \cite{icra24}    & Diffusion  & 3D Point Cloud  & \xmark \\
Zhang et al. \cite{ral24}      & Diffusion  & 2D BEV          & \cmark \\
\rowcolor{gray!15}
\name\ (Ours)            & Latent Diffusion & 2D BEV      & \cmark \\
\bottomrule
\end{tabular}
\vspace{-0.1in}
\end{table}

\section{Our Approach}
We leverage the denoising and generative capabilities of diffusion models to address the cross-modality challenge of transforming sparse, noisy radar measurements into dense LiDAR-like representations.  
Radar data naturally contains noise and artifacts, making it well matched to diffusion’s iterative denoising process.  
By conditioning the generation on radar inputs, we can guide the model to recover fine geometric detail while remaining faithful to the observed scene.  
This section describes our \name pipeline, starting from the fundamentals of diffusion models, moving to the radar input representation, and finally the key architectural choices that enable single-frame radar-to-LiDAR translation.  

\subsection{Primer on Diffusion Models}
\subsubsection{Denoising Diffusion Models}
\label{ddpm-priors}
Diffusion models are probabilistic generative models that learn to reverse a noise corruption process \cite{ddpm}.  
In our setting, the model learns to transform sparse radar inputs into high-resolution, LiDAR-like bird’s-eye view (BEV) images.  
\new

  
The framework has two stages: a fixed forward process $q$ and a learned reverse process $p_\theta$.  
The forward process gradually corrupts a clean LiDAR BEV $\mathbf{x}_0$ with Gaussian noise over $T$ timesteps:  
\begin{equation}
q(\mathbf{x}_t | \mathbf{x}_{t-1}) = \mathcal{N}(\mathbf{x}_t; \sqrt{1 - \beta_t}\mathbf{x}_{t-1}, \beta_t\mathbf{I}),
\end{equation}
where $\beta_t$ controls the variance at each step.  
As $t \rightarrow T$, the distribution converges to pure noise, erasing all structure from the original LiDAR image.  
\new

The reverse process $p_\theta$, parameterized by a U-Net, learns to invert this corruption.  
Starting from pure noise $\mathbf{x}_T \sim \mathcal{N}(\mathbf{0}, \mathbf{I})$, the model iteratively predicts and removes noise to reconstruct $\mathbf{\hat{x}}_0$.  
Conditioning on radar representation $\mathbf{c}$ guides the denoising so that the output reflects the true scene.  
Rather than directly predicting clean data, the network predicts the noise $\boldsymbol{\epsilon}_\theta(\cdot)$ added in the forward process.  
\new

Training minimizes the mean squared error between the predicted and true noise:  
\begin{equation}
L(\theta) = \mathbb{E}_{t, \mathbf{x}_0, \boldsymbol{\epsilon}_t} 
\left[ \left\| \boldsymbol{\epsilon}_t - \boldsymbol{\epsilon}_\theta(\mathbf{x}_t, t, \mathbf{c}) \right\|^2 \right],
\end{equation}
where $\mathbf{x}_t$ is sampled from the forward process.  
This objective anchors the denoising trajectory so that the final output is both geometrically sharp and consistent with the radar input.  
\new

\subsubsection{Latent Diffusion Models (LDM)}
Standard diffusion models operate directly in pixel space, requiring significant computational resources that scale poorly with image resolution. Latent diffusion models \cite{sd} address this by performing diffusion in a learned latent space. A pretrained VAE with encoder $\mathcal{E}$ and decoder $\mathcal{D}$ compresses images to latent representations, reducing computational cost from $O(H\times W)$ to $O(h\times w)$ where $h,w \ll H,W$.
\begin{equation}
    z = \mathcal{E}(x), \quad \hat{x} = \mathcal{D}(z)
\end{equation}
The diffusion framework now operates on latent vectors $\mathbf{z}$ instead of raw pixels. The forward process gradually adds noise to the clean latent $\mathbf{z_0}$ over $T$ timesteps, and the reverse process learns to predict the additive noise and denoise it from $\mathbf{z}_T \sim \mathcal{N}(\mathbf{0}, \mathbf{I})$. The training objective is adapted accordingly
\begin{equation}
L_{LDM}(\theta) = \mathbb{E}_{t, \mathcal{E}(\mathbf{x}_0), \boldsymbol{\epsilon}} \left[ \left\| \boldsymbol{\epsilon} - \boldsymbol{\epsilon}_\theta(\mathbf{z}_t, t, \mathbf{c}) \right\|^2 \right]
\end{equation}
where $\mathbf{z_t}$ is the noisy LiDAR latent representation at timestep $t$, $\mathbf{c}$ is the radar latent condition. Once the iterative denoising is finished, the decoder $\mathcal{D}$ maps the final, clean latent vector $\hat{\mathbf{z_0}}$ back into high-dimensional pixel space to produce the high-resolution LiDAR-like BEV $\mathbf{x_0}'=\mathcal{D}(\hat{\mathbf{z_0}})$.

\subsection{Radar Input Representation}
Our radar signal processing transforms raw I/Q signals into information-rich and structured representations compatible with diffusion models. We first apply range FFT and azimuth FFT to raw I/Q samples, generating raw 2D range-azimuth BEV images. Following RadarHD's preprocessing justification, we apply only light static thresholding (5\% of the magnitude) to preserve information from radar sidelobes and multipath reflections that typically camouflage as noise but contain valuable structural information. This preprocessing strategy integrates effectively with diffusion's denoising capabilities. Traditional algorithms discard weak signals and artifacts as noise, but these components often carry environmental cues about scene geometry. By providing rich, lightly-filtered heatmaps as input, we enable the diffusion model to distinguish salient features from true noise through learned priors.

\begin{figure}[tb]
    \centering
    \vspace{0.1in}
    \includegraphics[width=0.9\linewidth]{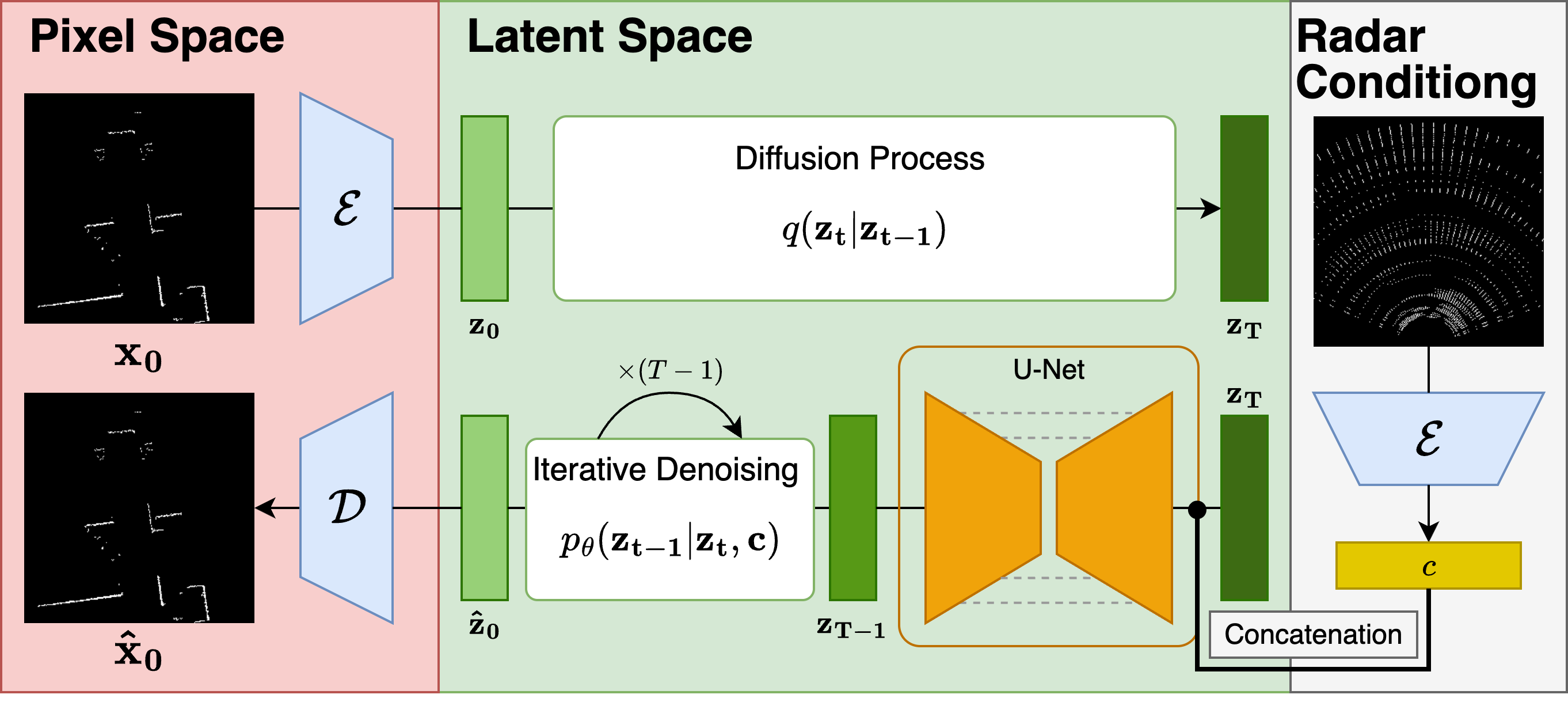}
    \caption{Overview of the \name's diffusion architecture. The radar and LiDAR images are first encoded into latent space. The radar latent $\mathbf{c}$ is concatenated with the noisy LiDAR latent $\mathbf{z}_t$ and fed into a pretrained U-Net denoiser. The U-Net iteratively removes noise while preserving structure as guided by the radar condition. After denoising, the decoder reconstructs a LiDAR-like point cloud with sharp geometry from a single radar frame.}
    \vspace{-0.2in}
    \label{fig-pipeline}
\end{figure}

\subsection{LDM Design Choices}
Our conditional latent diffusion model is built to recover fine-grained LiDAR-like structure from a single radar frame.  
Figure~\ref{fig-pipeline} illustrates the overall architecture: radar and LiDAR BEVs are first encoded by the frozen VAE into latent space, concatenated, and then denoised by a pretrained U-Net.  
Here we explain four key design aspects that enable this pipeline: the VAE architecture, the choice of pretrained U-Net backbone, the conditioning strategy, and the training objective.  
\new

\subsubsection{VAE Architecture}
We use the distilled Tiny AutoEncoder for Stable Diffusion (TAESD), which compresses inputs by a factor of $8\times$ while preserving fine detail.  
This allows the diffusion process to operate on compact latent representations ($4 \times 32 \times 64$ for our BEVs) rather than high-dimensional images, reducing computation by orders of magnitude.  
The VAE remains frozen during both training and inference. This provides three benefits:  
(1) it leverages robust pretrained image priors learned at web scale,  
(2) it ensures consistent latent properties for both radar and LiDAR BEVs, and  
(3) it keeps inference efficient with negligible added cost compared to a full VAE.  
Encoding both modalities with the same frozen VAE ensures their latents are directly compatible, enabling simple channel-wise concatenation for conditioning.  
\new

\subsubsection{Choice of Pre-trained U-Nets}
Training a diffusion backbone from scratch is infeasible for radar–LiDAR translation due to limited dataset size and high compute demands.  
Instead, we initialize from \textit{Marigold} \cite{marigold}, a state-of-the-art monocular depth estimator.  
Marigold’s U-Net has been optimized to infer scene depth and geometry from RGB images, which transfers well to our task of mapping sparse radar returns to dense LiDAR structure.  
We hypothesize that these “shape-preserving” priors provide a stronger starting point than training on radar–LiDAR pairs alone.  
For comparison, we also explore Stable Diffusion v2 (SDv2), which encodes broad semantic priors but less geometric bias, in our ablation studies.  
Our results confirm that depth-focused priors from Marigold provide superior reconstructions, while SDv2 offers a useful baseline for analyzing generalization.  
\new

\subsubsection{Conditioning Strategy}
A central question is how to inject radar information into the denoising U-Net.  
We explore and study two strategies inspired by Marigold and SDv2:  

\begin{itemize}
    \item \textbf{Channel-wise concatenation (Marigold-style).}  
    The radar BEV is encoded by the frozen VAE into latent $\mathbf{c}$, which is concatenated with the noisy LiDAR latent $\mathbf{z}_t$ along the channel dimension.  
    This provides direct spatial alignment, anchoring diffusion to the observed radar geometry and preserving structural fidelity.  
    \item \textbf{Cross-attention (SDv2-style).}  
    The radar BEV is first projected into a sequence of embeddings, which are then injected into the U-Net at each layer through cross-attention.  
    This provides more flexibility, allowing the model to learn higher-level correlations across modalities.  
\end{itemize}

We find that each strategy presents a trade-off. Concatenation is simple and provides strong geometric guidance but may limit generalization.  
Cross-attention enables more abstract relationships but loses explicit spatial alignment and requires training an additional encoder.  
In this work, we adopt concatenation as the primary conditioning scheme, and evaluate cross-attention in ablation studies to quantify the trade-off.  
\new

\subsubsection{Training Objective}
A standard LDM is trained with a latent noise prediction loss Eq. 4, which encourages the predicted noise to match the true Gaussian noise added during the forward process.  
While this ensures distributional consistency, it does not guarantee structural accuracy after decoding.  
As a result, the model may generate “LiDAR-like” outputs that look plausible but correspond to the wrong scene.  

To address this, we add a pixel-space reconstruction loss $L_p$ to tether generations to the true LiDAR structure:  
\begin{equation}
L_p = \lambda_{L1} L_{L1} + \lambda_{SSIM} L_{SSIM} + \lambda_{LPIPS} L_{LPIPS}.
\end{equation}
Here, $L_{L1}$ is the Mean absolute error that enforces per-pixel accuracy, $L_{SSIM}$ is the loss from SSIM that preserves structural integrity, and $L_{LPIPS}$ is the LPIPS loss that aligns perceptual similarity with human judgments.  

The final training objective is a weighted combination of latent and pixel-space losses:  
\begin{equation}
L_{total} = L_{LDM} + \lambda_p L_p.
\end{equation}

This dual-space objective balances distribution matching with per-scene fidelity.  
It reduces hallucinations, enforces sharper geometry, and ensures reconstructions remain consistent with the radar input. 
\new

\section{Evaluation}
We evaluate \name\ on the RadarHD dataset, comparing against traditional and learning-based baselines.  
Our analysis covers dataset setup, evaluation metrics, baseline descriptions, and point cloud comparisons.  
\new

\subsection{Dataset}
Most publicly available radar datasets target automotive applications with outdoor environments \cite{barnes2020oxford,caesar2020nuscenes,radelft}. Indoor-focused datasets like RaDiCal \cite{lim2021radical} and RadarRGBD \cite{song2025radarrgbd} provide raw I/Q signals but lack corresponding LiDAR ground truth. View-of-Delft (VoD) \cite{vod} offers radar-LiDAR pairs but only provides processed radar point clouds rather than raw signals, limiting preprocessing flexibility. Additionally, VoD focuses on outdoor driving scenarios.

ColoRadar \cite{kramer2022coloradar} comes closest to our requirements, providing both raw radar signals and LiDAR ground truth. However, indoor samples comprise only 15k of the dataset. RadarHD dataset addresses this gap with a larger collection of indoor radar-LiDAR pairs captured using single-chip mmWave systems. The dataset contains approximately 40k data pairs of raw mmWave radar signals with high-resolution LiDAR ground truth collected across 67 diverse trajectories indoor and outdoor. Given its primary focus on indoor scenarios and availability of raw radar representations, we adopt RadarHD as our training and evaluation dataset.

Evaluation is performed on the official test split, which contains trajectories not used during training.  
The split is designed to measure generalization under three conditions: (i) New trajectories in the same environment, (ii) trajectories from similar indoor environments, and (iii) trajectories from unseen and visually distinct environments.  
\new

\begin{table}[tb]
\centering
\vspace{0.1in}
\caption{2D Radar Super-Resolution Performance}
\label{tab-results}
\setlength{\tabcolsep}{6pt}
\renewcommand{\arraystretch}{1.25}
\begin{tabular}{L{0.35\linewidth} C{0.1\linewidth} C{0.1\linewidth} C{0.1\linewidth}}
\toprule
\textbf{Work} & \textbf{\# of Frames} & \textbf{Mean CD$\downarrow$} & \textbf{Mean MHD$\downarrow$} \\
\midrule
CFAR & 1 & 0.84 & 0.91 \\
RadarHD \cite{radarhd} & 41 & 0.44 & 0.34 \\
RadarHD \cite{radarhd} single-frame & 1 & 0.56 & 0.45 \\
Luan et al. \cite{icra24} & 5 & 0.59 & 0.50 \\
Zhang et al. \cite{ral24} & 1 & 0.38 & 0.29 \\
\rowcolor{gray!15}
\textbf{\name\ (Ours)} & 1 & \textbf{0.35} & \textbf{0.28} \\
\bottomrule
\end{tabular}
\vspace{-0.2in}
\end{table}

\subsection{Metrics}
We extract 2D point clouds from both the generated BEVs and the corresponding LiDAR ground truth, and compute two widely used similarity metrics: (i) \textbf{Chamfer Distance (CD)}: the average bidirectional distance between each point and its nearest neighbor in the other point cloud. (ii) \textbf{Modified Hausdorff Distance (MHD)}: the mean nearest-neighbor distance, more robust to outliers than the classical Hausdorff distance. Lower values indicate closer agreement with the LiDAR reference.  
\new

\subsection{Baselines}
We compare \name\ against both traditional and recent learning-based methods:  
\begin{itemize}
    \item \textbf{CFAR} \cite{cacfar}: A conventional CA-CFAR detector with 5\,dB threshold, representing non-learning approaches.  
    \item \textbf{RadarHD} \cite{radarhd}: A VAE-based model trained with 41 stacked radar frames (temporal SAR).  
    \item \textbf{RadarHD (single-frame)}: The same model evaluated with only one frame repeated 41 times, to measure performance without SAR cues.  
    \item \textbf{Luan et al.} \cite{icra24}: A diffusion-based method using 5 temporal frames, primarily targeting 3D point cloud generation.  
    \item \textbf{Zhang et al.} \cite{ral24}: A recent diffusion framework designed for single-frame inference.  
\end{itemize}

\begin{figure}[b]
\vspace{-0.1in}
    \centering
    \includegraphics[width=0.48\linewidth]{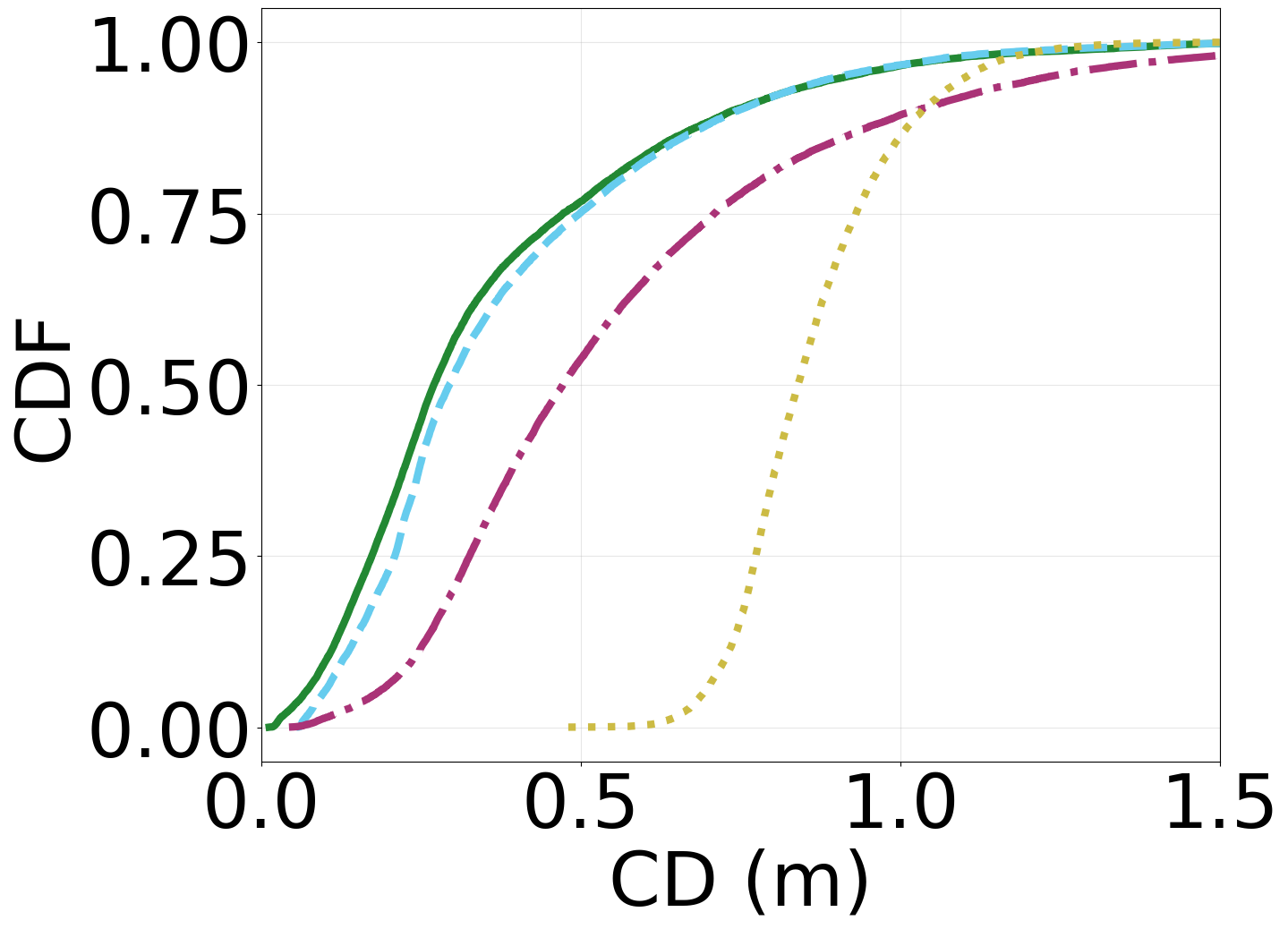}
    \includegraphics[width=0.48\linewidth]{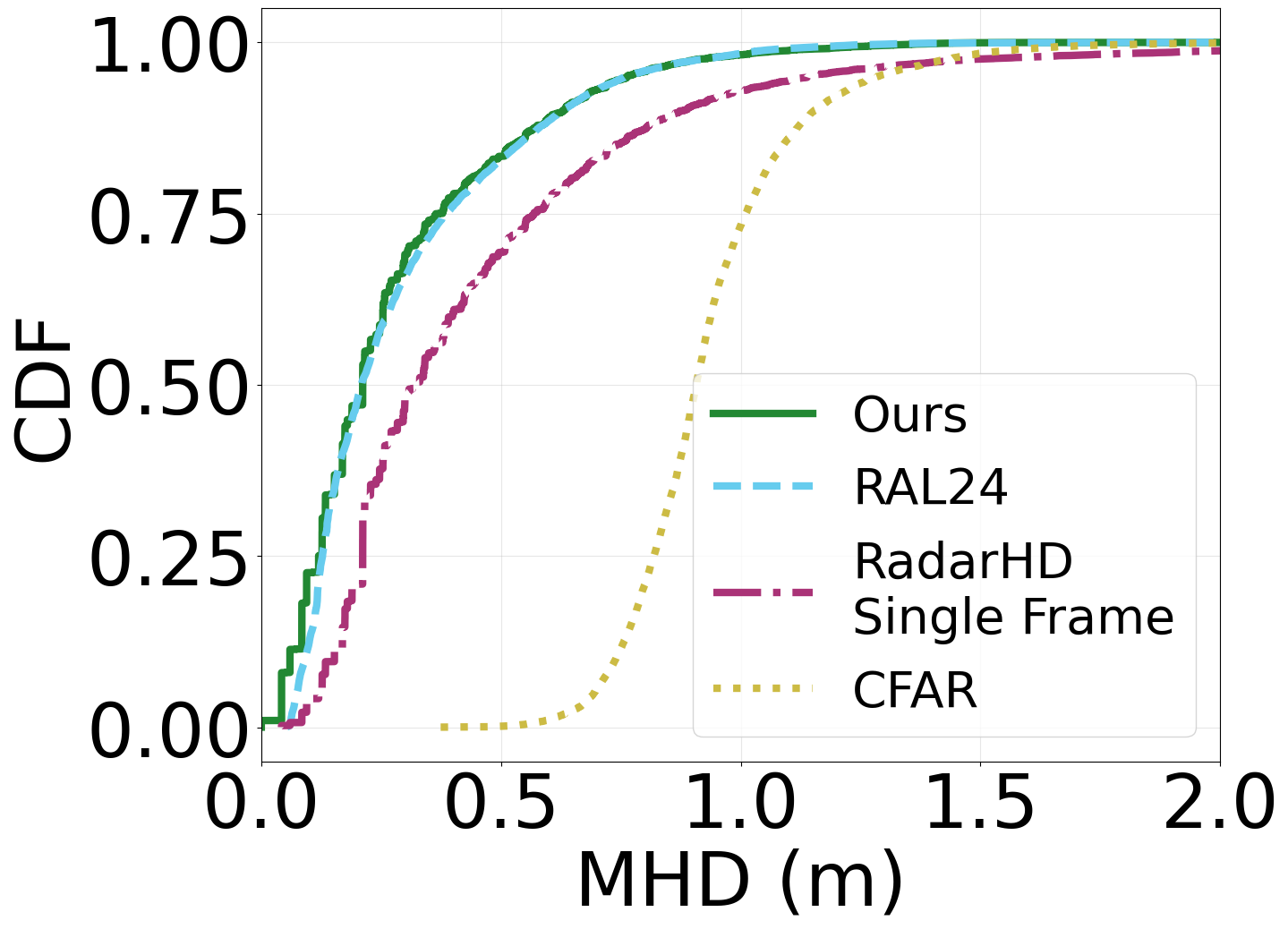}
    \caption{CDFs of reconstruction error (CD, MHD). \name\ achieves the lowest errors, outperforming Zhang et al., RadarHD single-frame, and CFAR, with most samples under 0.5\,m CD and 0.4\,m MHD.}
    \vspace{-0.05in}
    \label{fig:cdf-combined}
\end{figure}

\begin{figure*}[t] 
    \vspace{0.1in}
    \centering 
    \setlength{\tabcolsep}{0pt}
    \renewcommand{\arraystretch}{1.0} 
    
    \begin{tabular}{|c|c|c|c|c|c|}
        \hline 
        \includegraphics[width=0.14\textwidth, valign=m]{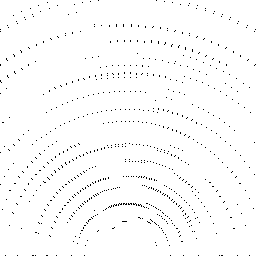} &
        \includegraphics[width=0.14\textwidth, valign=m]{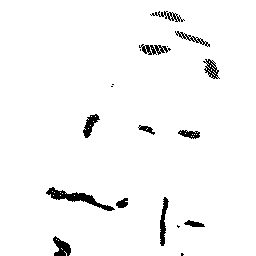} &
        \includegraphics[width=0.14\textwidth, valign=m]{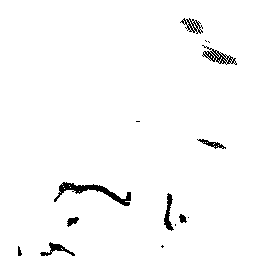} &
        \includegraphics[width=0.14\textwidth, valign=m]{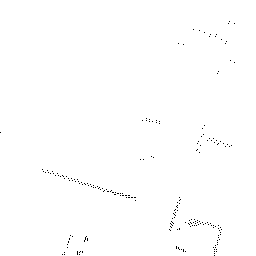} &
        \includegraphics[width=0.14\textwidth, valign=m]{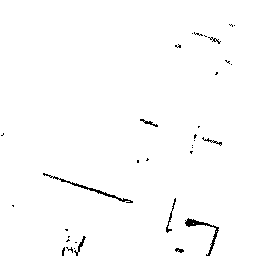} &
        \includegraphics[width=0.14\textwidth, valign=m]{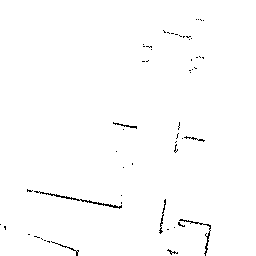} \\
        \hline 
        \includegraphics[width=0.14\textwidth, valign=m]{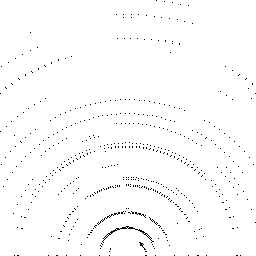} &
        \includegraphics[width=0.14\textwidth, valign=m]{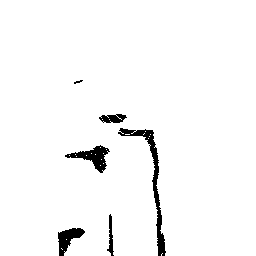} &
        \includegraphics[width=0.14\textwidth, valign=m]{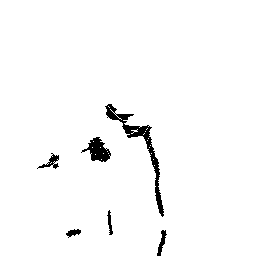} &
        \includegraphics[width=0.14\textwidth, valign=m]{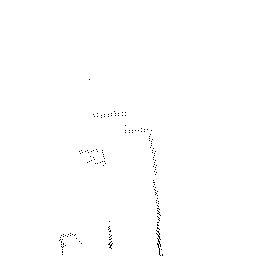} &
        \includegraphics[width=0.14\textwidth, valign=m]{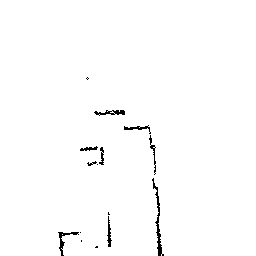} &
        \includegraphics[width=0.14\textwidth, valign=m]{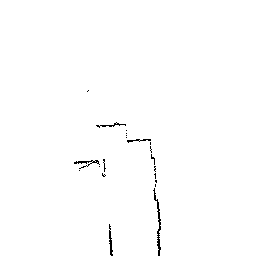} \\
        \hline 
        \includegraphics[width=0.14\textwidth, valign=m]{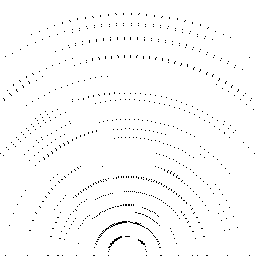} &
        \includegraphics[width=0.14\textwidth, valign=m]{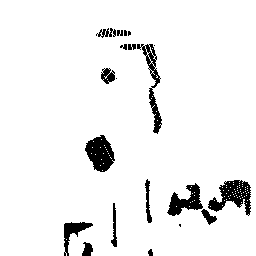} &
        \includegraphics[width=0.14\textwidth, valign=m]{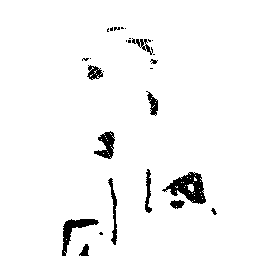} &
        \includegraphics[width=0.14\textwidth, valign=m]{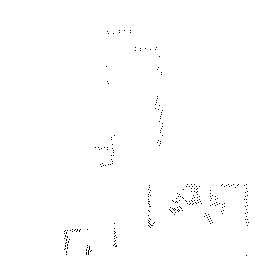} &
        \includegraphics[width=0.14\textwidth, valign=m]{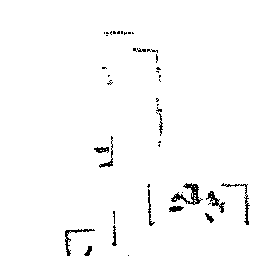} &
        \includegraphics[width=0.14\textwidth, valign=m]{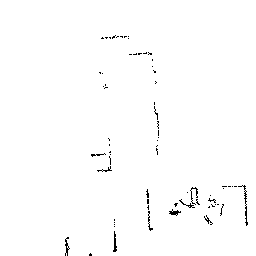} \\
        \hline 
        \includegraphics[width=0.14\textwidth, valign=m]{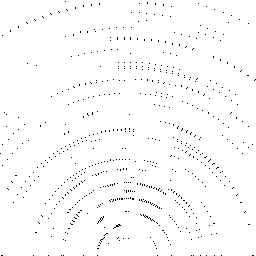} &
        \includegraphics[width=0.14\textwidth, valign=m]{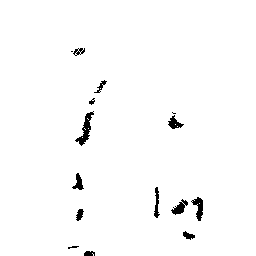} &
        \includegraphics[width=0.14\textwidth, valign=m]{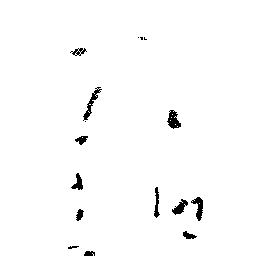} &
        \includegraphics[width=0.14\textwidth, valign=m]{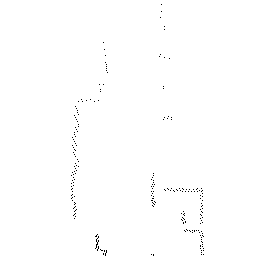} &
        \includegraphics[width=0.14\textwidth, valign=m]{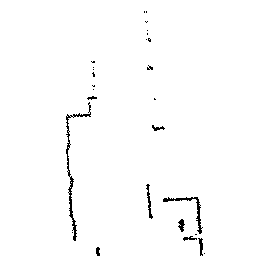} &
        \includegraphics[width=0.14\textwidth, valign=m]{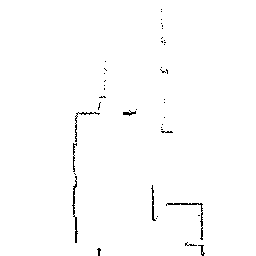} \\
        \hline
        \multicolumn{1}{c}{(a) CA-CFAR} &
        \multicolumn{1}{c}{(b) RadarHD} &
        \multicolumn{1}{c}{(c) RadarHD single-frame} &
        \multicolumn{1}{c}{(d) Zhang et al.} &
        \multicolumn{1}{c}{(e) Ours (\name)} &
        \multicolumn{1}{c}{(f) LiDAR ground truth} \\
    \end{tabular}
    \caption{Qualitative comparison of point cloud reconstructions on four representative scenes with varying complexity. CA-CFAR is sparse and noisy; RadarHD \cite{radarhd} (41 frames) captures geometry but with blurry edges and clutter; RadarHD single-frame misses structures; Zhang et al. \cite{ral24} yields cleaner but low-resolution (128×128) outputs. Our method achieves sharp, complete reconstructions at 256×512, closely matching ground-truth LiDAR and showing clearer wall boundaries with fewer hallucinations. All results are shown in Cartesian coordinates for direct comparison.}
    \label{fig-qual}
    \vspace{-0.1in}
\end{figure*}

\subsection{Point Cloud Comparison}
Table \ref{tab-results} summarizes the mean CD and MHD across methods.
As expected, CFAR has the highest error due to sparse detections.  
RadarHD achieves strong results with 41 stacked frames but degrades sharply in the single-frame case (0.56\,m CD, 0.45\,m MHD), underscoring its dependence on SAR.  
Both diffusion-based methods improve in the single-frame setting: Zhang et al. achieves 0.38\,m CD and 0.29\,m MHD, while \name\ achieves 0.35\,m CD and 0.28\,m MHD, a reduction of 8\% and 3\% respectively.  
At first glance, the numbers suggest parity, but the two methods differ fundamentally in design.  
Zhang et al. operates directly in pixel space, which makes inference slower (2.4\,s per frame) and training less data-efficient.  
In contrast, \name\ performs diffusion in latent space, compressing inputs through a frozen VAE.  
This reduces inference to 1.3\,s per frame, lowers compute requirements, and enables transfer of pretrained priors for generalization to unseen environments (shown in Fig. \ref{fig-generalization})

CDF plots for CD and MHD in Figure~\ref{fig:cdf-combined} illustrate the error distribution across test samples.  
Our method consistently shifts the curves leftward compared to baselines, indicating lower errors across the dataset.  
Notably, the \name\ curve overtakes the Zhang et al. baseline \cite{ral24}, showing that the majority of scenes benefit from our latent-diffusion approach.  
This effect is especially clear in the MHD plot, where \name\ achieves lower tail errors, highlighting improved robustness to outliers.  
These results confirm that transferring pretrained depth priors into the radar-to-LiDAR pipeline leads to more faithful reconstructions.  
\new

Figure~\ref{fig-qual} provides qualitative comparison of reconstruction across baselines.  
We exclude Luan et al. \cite{icra24} since its implementation is not public and its reported visualizations focus on the VoD dataset.  
RadarHD with 41 temporal frames captures coarse scene contours but often produces blurred boundaries and misses fine structural detail.  
For example, in the first row, RadarHD introduces clutter near the scene center, while its single-frame version drops structures near the lower right region.  
These artifacts reflect the limitations of its VAE backbone, which lacks the denoising strength of diffusion models.  
By contrast, diffusion-based methods generate sharper and more complete geometry.  
\new

Both Zhang et al. \cite{ral24} and \name\ recover the overall shape of the scene.  
However, Zhang et al. operates on downsampled $128 \times 128$ BEVs, producing coarser details, while \name\ reconstructs directly at $256 \times 512$ resolution, yielding finer structural fidelity.  
Qualitatively, this is visible in sharper wall boundaries and cleaner gaps in our reconstructions.  
\new

\begin{table*}[t]
\vspace{0.05in}
\centering
\caption{Ablation study of \name on RadarHD dataset.}
\label{tab-ablation}
\setlength{\tabcolsep}{6pt}
\renewcommand{\arraystretch}{1.2}
\begin{tabular}{l l l l c c}
\toprule
\textbf{Experiment} & \textbf{Radar Input} & \textbf{Backbone} & \textbf{Conditioning} & \textbf{Mean CD [m]$\downarrow$} & \textbf{Mean MHD [m]$\downarrow$} \\
\midrule
Zero-threshold BEV      & BEV (zero-thresh) & Marigold (pretrained) & Concatenation & 0.36 & 0.29 \\
Raw I/Q input           & I/Q signals     & Marigold (pretrained) & Cross-attn     & \textit{0.81} & \textit{0.74} \\
Random init             & BEV (light-thresh)& Marigold (random)     & Concatenation & \textit{0.85} & \textit{1.08} \\
Alt. pretraining (SDv2) & BEV (light-thresh)& SDv2 (pretrained)     & Cross-attn     & 0.39 & 0.32 \\
Latent-only loss        & BEV (light-thresh)& Marigold (pretrained) & Concatenation & \textit{1.12} & \textit{1.12} \\
\midrule
L1 only                 & BEV (light-thresh)& Marigold (pretrained) & Concatenation & 0.35 & 0.28 \\
L1 + SSIM               & BEV (light-thresh)& Marigold (pretrained) & Concatenation & 0.42 & 0.34 \\
L1 + LPIPS              & BEV (light-thresh)& Marigold (pretrained) & Concatenation & 0.35 & 0.28 \\
\midrule
\rowcolor{gray!15}
\textbf{\name} & BEV (light-thresh) & Marigold (pretrained) & Concatenation & \textbf{0.35} & \textbf{0.28} \\
\bottomrule
\end{tabular}
\vspace{-0.1in}
\end{table*}

\begin{figure*}[htb]
    \centering
    \includegraphics[width=0.32\linewidth]{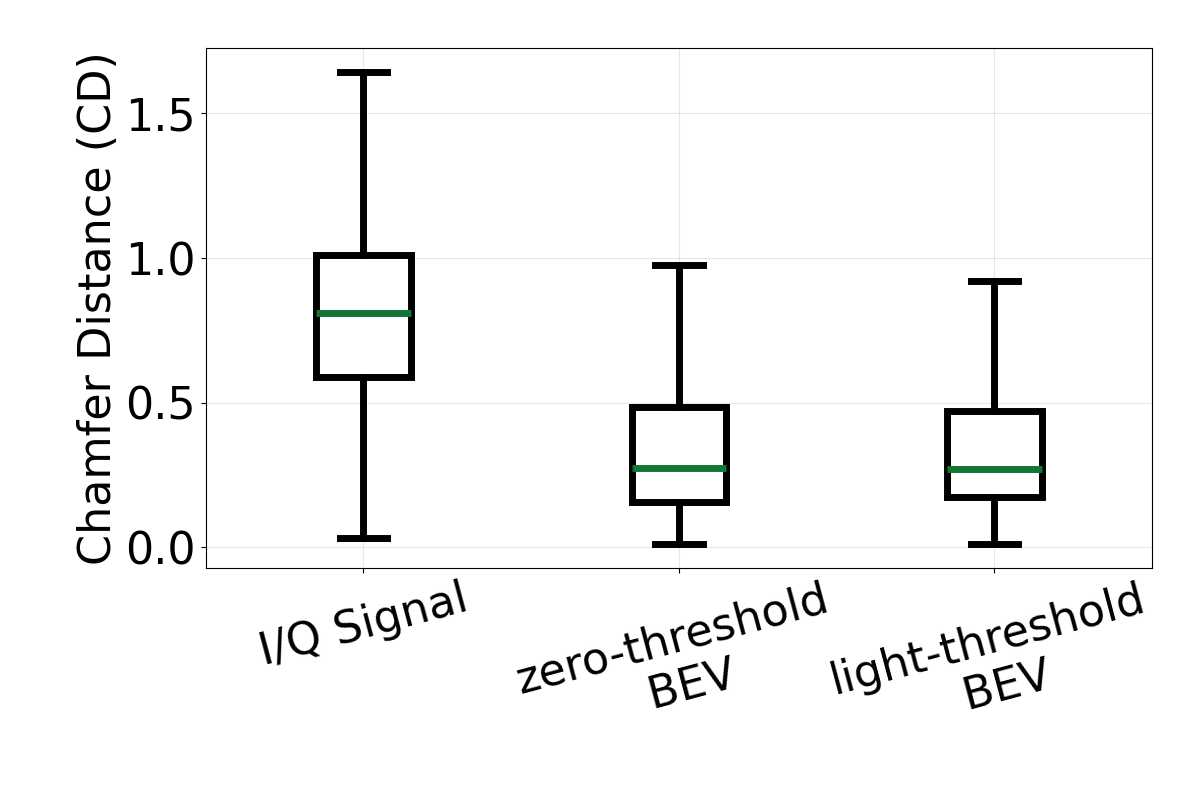}\hfill
    \includegraphics[width=0.32\linewidth]{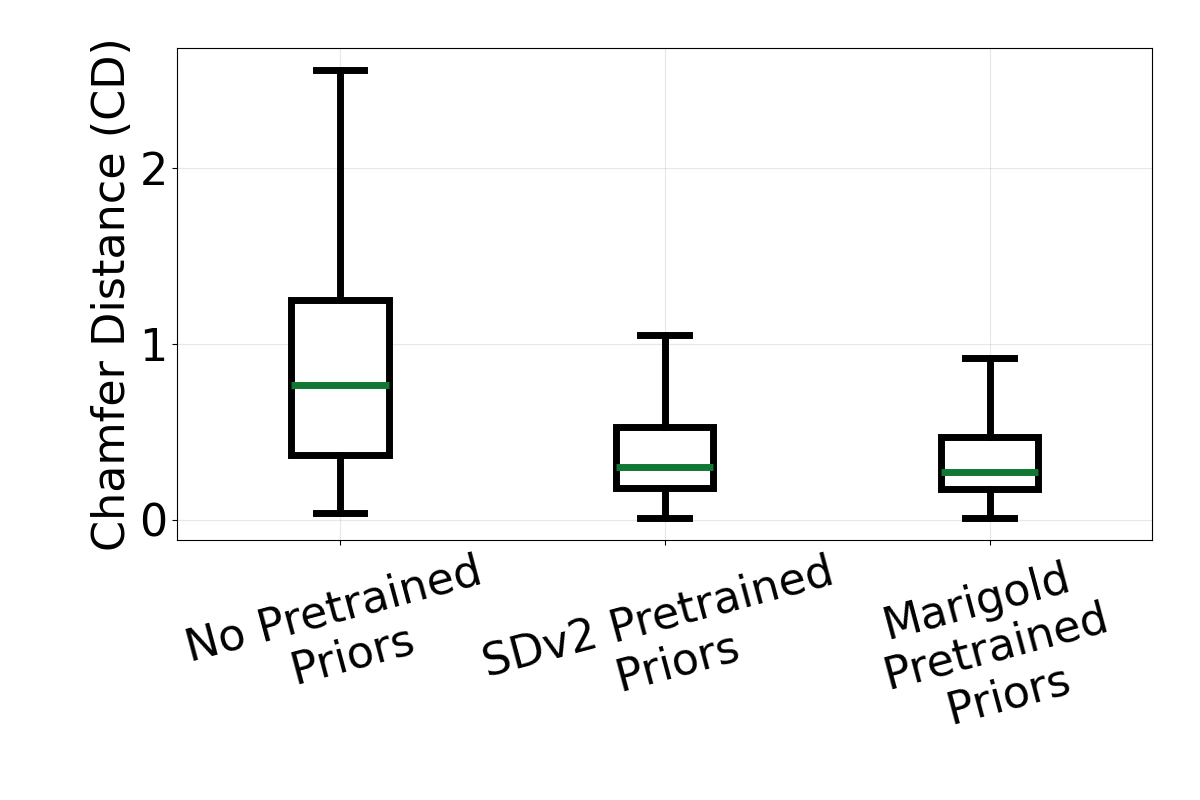}\hfill
    \includegraphics[width=0.32\linewidth]{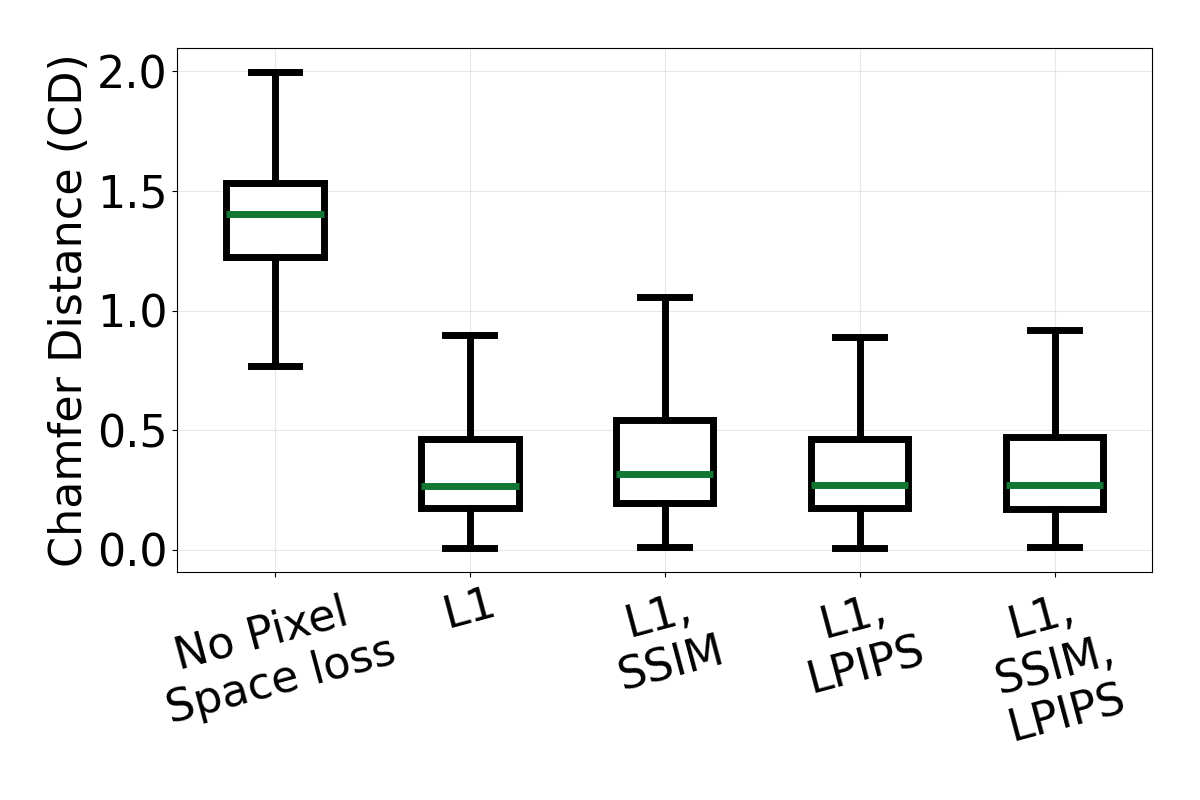}
    \vspace{-0.1in}
    \caption{Ablation box plots (Chamfer Distance). Left to right: input representation, pretrained priors, training losses. Thresholded BEV inputs outperform raw I/Q; depth-pretrained priors (Marigold, SDv2) beat random initialization; adding pixel-space L1 drives most of the gain, with SSIM and LPIPS providing only marginal benefit.}
    \vspace{-0.2in}
    \label{fig:ablation-full}
\end{figure*}

Diffusion models can also introduce artifacts.  
In the second row, both Zhang et al. and \name\ produce hallucinated points near the lower-left region.  
Such failure cases - minor hallucinations or occasional missing geometry - likely stem from the limited scale of the RadarHD dataset.  
These observations suggest that larger and more diverse training corpora would further improve the reliability of single-frame radar-to-LiDAR translation.  
\new

\subsection{Ablation Study}
We evaluate the impact of key design choices in \name: radar input representation, pretrained priors with conditioning, and training objectives. Results are summarized in Table~\ref{tab-ablation} and Figure~\ref{fig:ablation-full}.  
\new

\subsubsection{Radar Input Representation}
We compare light-threshold BEV (our default), zero-threshold BEV, and raw I/Q inputs.  
Zero-threshold BEV performs nearly on par with light-threshold, showing that mild noise filtering is sufficient for diffusion models.  
In contrast, directly using raw I/Q signals with a custom CNN encoder increases Chamfer Distance by $2.3\times$, reflecting the difficulty of learning radar physics from scratch without priors optimized for image-like inputs.  
This confirms that lightly preprocessed BEV heatmaps are the most effective representation for conditioning.  
\new

\subsubsection{Pretrained Priors and Conditioning}
Next we evaluate the role of pretrained weights and conditioning strategies.  
Without pretrained priors, performance degrades by nearly 3$\times$ and variance increases sharply, with errors exceeding 2.5\,m.  
This highlights that geometric priors learned by Marigold’s depth estimation training are essential for cross-modal translation.  
Initializing from Stable Diffusion v2 yields reasonable results but underperforms Marigold, validating that task-specific geometric priors outperform generic semantic ones.  

We also tested radar conditioning via cross-attention, as in SDv2. While flexible, cross-attention underperforms concatenation across both I/Q inputs and SDv2 priors.  
Channel-wise concatenation provides explicit spatial alignment between radar and LiDAR latents, leading to sharper reconstructions and stronger geometric fidelity.  
\new

\subsubsection{Training Objectives}
Finally, we analyze training objectives.  
Latent-only supervision performs worst, with errors exceeding 1\,m, confirming that latent alignment alone does not guarantee structural fidelity after decoding.  
Adding pixel-space losses dramatically improves performance.  
L1-only training already matches our full system, while combinations with SSIM or LPIPS provide complementary benefits.  
Our complete formulation (L1+SSIM+LPIPS) offers the best consistency across scenes, balancing numerical accuracy, structural similarity, and perceptual fidelity.  
\new

\section{Conclusion}
We introduce \name, a latent diffusion model that reconstructs dense LiDAR-like point clouds from single mmWave radar frames without requiring SAR.  
By transferring pretrained depth priors into latent diffusion, \name\ achieves state-of-the-art accuracy while remaining efficient and practical for SWaP-constrained robotic platforms. 
Our results and ablations highlight the importance of pretrained priors, BEV conditioning, and dual-space objectives, offering a general recipe for cross-modal sensor translation.

\bibliographystyle{IEEEtran}
\bibliography{ref}

\end{document}